\documentclass{article}
\usepackage{amssymb}
\usepackage{subcaption}
\usepackage{amsmath}

 \usepackage[final]{corl_2025} 

\usepackage{graphicx}
\usepackage{svg}
\usepackage{tabularx}
\usepackage{CJKutf8}
\usepackage{footnote}

\title{Mechanistic Interpretability for Steering Vision-Language-Action Models}

\author{
  Bear H\"{a}on\thanks{Equal contribution. Correspondence to: \texttt{\{bear.haon, kaylene\}@berkeley.edu}} \quad
  Kaylene Stocking\footnotemark[1]\quad
  Ian Chuang\quad
  Claire Tomlin\\
  Department of Electrical Engineering and Computer Sciences \\
  University of California, Berkeley
}

\begin{document}
\maketitle

\makeatletter
\renewcommand\@makefnmark{}
\makeatother

\begin{figure}[htbp]
  \vspace{-15pt} 
  \centering
  \includegraphics[width=1.0\textwidth]{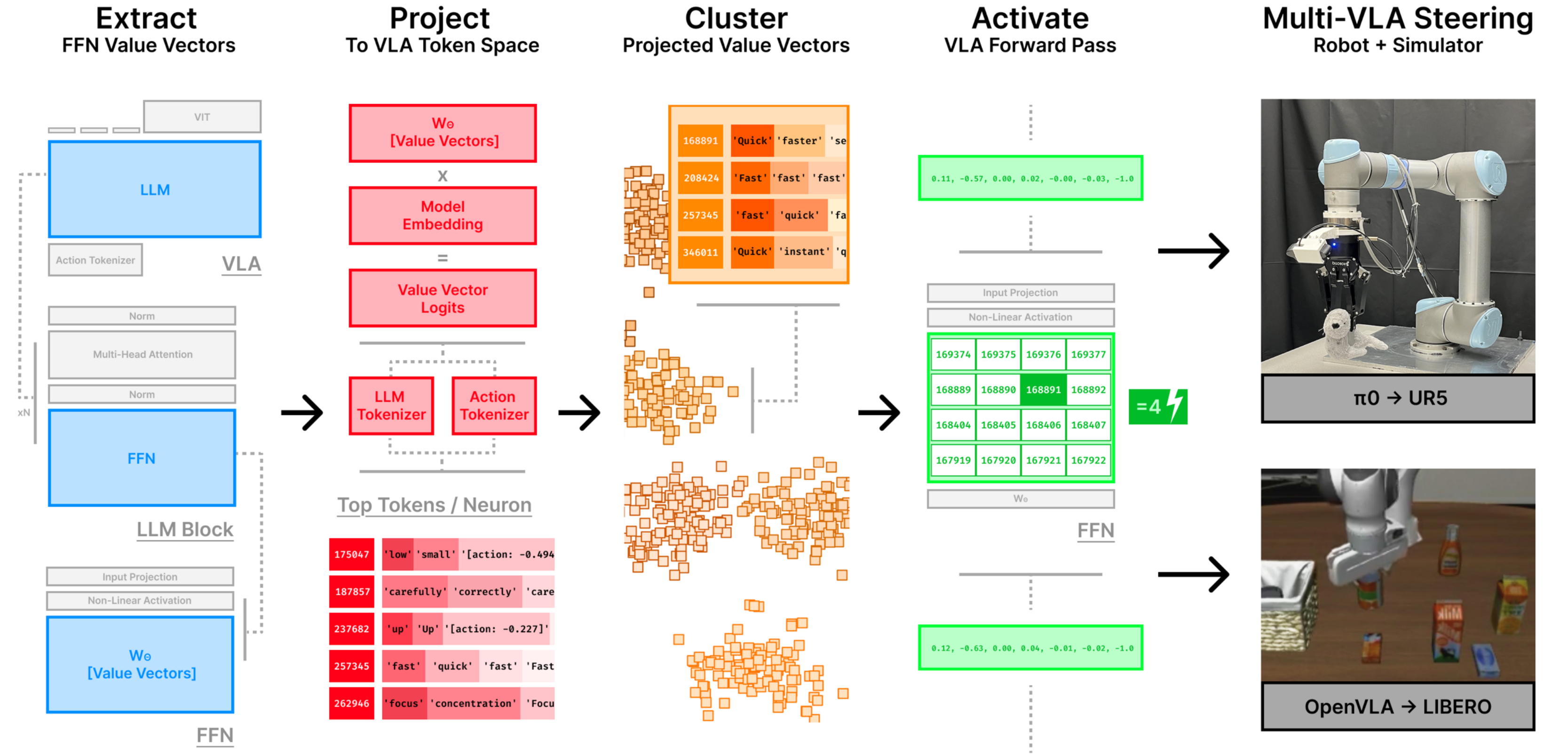} 
    \caption{\textbf{We present a framework for steering Vision-Language-Action (VLA) models.} We extract FFN vectors, project them to the VLA token space, cluster them by semantic alignment, and inject activations at inference time to modulate behavior. Our experiments demonstrate interpretable zero-shot control in both simulation (\textsc{OpenVLA} in LIBERO) and on a physical robot ($\pi_0$ on a UR5).}

  \label{fig:main_graphic}
  \vspace{-10pt}
\end{figure}


\begin{abstract}
Vision-Language-Action (VLA) models are a promising path to realizing generalist embodied agents that can quickly adapt to new tasks, modalities, and environments. However, methods for interpreting and steering VLAs fall far short of classical robotics pipelines, which are grounded in explicit models of kinematics, dynamics, and control. This lack of mechanistic insight is a central challenge for deploying learned policies in real-world robotics, where robustness and explainability are critical. Motivated by advances in mechanistic interpretability for large language models, \textbf{we introduce the first framework for interpreting and steering VLAs via their internal representations, enabling direct intervention in model behavior at inference time.} We project feedforward activations within transformer layers onto the token embedding basis, identifying sparse semantic directions - such as \textit{speed} and \textit{direction} - that are causally linked to action selection. Leveraging these findings, we introduce a general-purpose activation steering method that modulates behavior in real time, without fine-tuning, reward signals, or environment interaction. We evaluate this method on two recent open-source VLAs, $\pi_0$ and \textsc{OpenVLA}, and demonstrate zero-shot behavioral control in simulation (LIBERO) and on a physical robot (UR5). \textbf{This work demonstrates that interpretable components of embodied VLAs can be systematically harnessed for control---establishing a new paradigm for transparent and steerable foundation models in robotics.}
\end{abstract}



\section{Introduction}

Large language-conditioned models represent a paradigm shift for robotics. By leveraging Internet-scale natural language and image data, these models promise to yield generalist robot policies that can follow directions, exhibit common-sense reasoning about their environments, and quickly adapt to novel tasks and scenarios. 

One leading type of frontier robotics model is the Vision-Language-Action (VLA) model, where a pre-trained Vision-Language Model (VLM) is trained to produce robot actions conditioned on a task description and image observations. While most VLAs are designed to be fine-tuned for specific  tasks, future models may even be able to perform a variety of tasks without the need for fine-tuning. Indeed, the $\pi_0$ model \cite{black2024pi0visionlanguageactionflowmodel} can often complete zero-shot evaluation tasks on the DROID platform \cite{pertsch2025fast} and  Gemini Robotics demonstrates zero-shot control via code generation \cite{geminiroboticsteam2025geminiroboticsbringingai}. This promise of generality and ease of deployment signals a large shift in how we deploy robot systems. In contrast to rigorous evaluation and certification on individual tasks in factories or warehouses, robots will be placed in novel environments and expected to behave well ``out-of-the-box''. This raises important questions about safety, robustness, and transparency. How and when can we trust that VLA policies will behave safely? When they fail, how can we diagnose and fix the underlying problems?

A similar set of questions and challenges is being faced by Large Language Model (LLM) researchers. While they have by no means been solved, the field of \emph{mechanistic interpretability} has yielded a number of exciting insights about the inner workings of LLMs. For example, researchers have discovered the functional role of specific attention heads \cite{olsson2022context} and identified semantically interpretable concepts in model activations \cite{bricken2023monosemanticity,cunningham2023sparse}. There is also evidence that mechanistic interpretability can be useful for diagnosing problems with models in practice \cite{marks2025auditing}. These insights lead us to ask: can mechanistic interpretability also help with understanding and shaping the behavior of frontier models in robotics? 

In this work, we present a series of experiments that together allow us to provide a resounding affirmative answer. We focus our efforts on an interpretability method that analyzes the semantic meanings of vectors in the output weight matrix of the transformer block's feed-forward layer (FFN) \cite{geva2022transformer}. Although this technique has less power to disentangle concepts that might be spread across multiple neurons in the model than alternatives such as sparse autoencoders \cite{bricken2023monosemanticity}, it allows us to interpret and steer the model without any additional training data. This is a key advantage in robotics where training data is expensive and difficult to acquire, and where available open datasets are small compared to what is available in the natural language domain. Our chosen method leads us to several surprising conclusions about VLAs, including:
\begin{enumerate}
    \item Despite being trained to produce only robot actions, the internal activity of the VLA model is largely semantic, with less than 25\% of FFN neurons being rewired for action prediction and a large fraction of the remainder representing clear, semantically interpretable concepts.
    \item Internal model concepts such as ``slow'' are causally linked to robot actions that express these concepts (i.e. moving the end effector more slowly), despite the model being trained without any feedback on the meaning or consequences of different action tokens.
    \item Targeted interventions on model concepts allow VLA behavior to be modified at inference time out-of-the-box, without any fine-tuning data or environment interaction required to select the intervention.
\end{enumerate}

These contributions represent the first attempt to use mechanistic interpretability techniques developed for LLMs with large robotics models. Our exciting initial results speak to the great promise of adapting these techniques to the robotics context, where there is an especially strong need for AI models that are safe and transparent.


\section{Related Work}
\label{sec:relatedwork}

\textbf{Mechanistic Interpretability and Robotics.} The size and complexity of deep learning systems make it challenging for humans to understand how they arrive at a particular output from a given input. To help address this problem, a variety of methods have been proposed to help expose and interpret the internal representations and computations in models. The most powerful form of interpretability work doesn't simply yield qualitative insights, but elucidates the causal mechanisms that determine the output of the model - this is often referred to as ``mechanistic interpretability''. Despite promising developments in mechanistic interpretability in the vision \cite{olah2017feature} and natural language \cite{bricken2023monosemanticity,lindsey2025biology,meng2022locating} settings, there has been relatively little work investigating DNN interpretability for robotics. In \cite{wang2023measuring}, they aim to interpret a deep learned policy by associating its factors of variation with logic rules, but the system they investigate is small in scale. In \cite{pohland2024understanding}, they use techniques inspired by vision interpretability work to improve a robotic image-condititioned policy. Finally, there has been recent interest in interpretability for reinforcement learning policies, as surveyed in \cite{glanois2024survey}.


\section{Interpreting VLAs}
\label{sec:interpreting}

Deep neural networks build their predictions layer by layer, with features in one layer supporting further computations and representations in the next. According to the ``linear representation hypothesis'', intermediate network layers often represent different features or concepts as directions in the latent embedding space \cite{park2024linear}. This means that if, for example, a model represents the concept of ``fast'', if we can identify the corresponding direction in its latent space, we can understand when and how the model uses this concept to produce its final output. Serendipitously, it turns out that for many large models, semantically meaningful directions align with individual neurons. This fact has been used to help interpret both vision \citep{olah2017feature} and language \citep{geva2021transformer,geva2022transformer} models. Recently, there has been excitement around using sparse autoencoders (SAEs) to uncover even more meaningful directions \cite{bricken2023monosemanticity}. However, training SAEs require large amounts of data, and the results vary depending on the exact dataset used. Both of these factors are significant barriers in the robotics context, where open-source datasets are limited and are unlikely to cover all of the semantic concepts that the model learned and might retain from VLM pretraining. Therefore, in our work, we focus on single-neuron features.

Specifically, we follow the approach outlined in \cite{geva2022transformer} and focus on the feedforward layer of each transformer block:
\begin{equation}
    \label{eqn:FFN}
    \text{FFN}(x) = f_\theta(x)^TW_\theta
    \vspace{1ex}
\end{equation}
where $x \in \mathbb{R}^n$ is the input to the layer, $f_\theta(x) \in \mathbb{R}^{m}$ is an input-dependent set of activations, and $W_\theta \in \mathbb{R}^{m \times n}$ is an input-independent parameter matrix. Using $w^{(i)}_\theta \in \mathbb{R}^n$ to refer to the $i$th row of $W_\theta$, Equation~(\ref{eqn:FFN}) can be rewritten as follows:
\begin{equation}
    \label{eqn:FFN_decomposition}
    \text{FFN}(x) = \sum_i[f_\theta(x)]_iw^{(i)}_\theta
\end{equation}
\vspace{-3ex}

This allows us to interpret the FFN as a weighted sum over the $w^{(i)}_\theta$ vectors, which we refer to as \emph{value vectors}. Since these vectors are independent of the inputs, they can be thought of as basis functions for the output of the FFN. Furthermore, as they belong to the same linear space as the final output of the transformer, they can be interpreted as probability distributions over possible output tokens. Therefore, we can assign semantic meanings to value vectors based on the set of tokens to which they assign the highest probability.

\subsection{How does VLA Training Affect Concepts Learned During VLM Pre-training?}

\begin{figure}[t]
    \centering
    \begin{subfigure}[t]{0.62\textwidth}
        \centering
        \includegraphics[width=\linewidth]{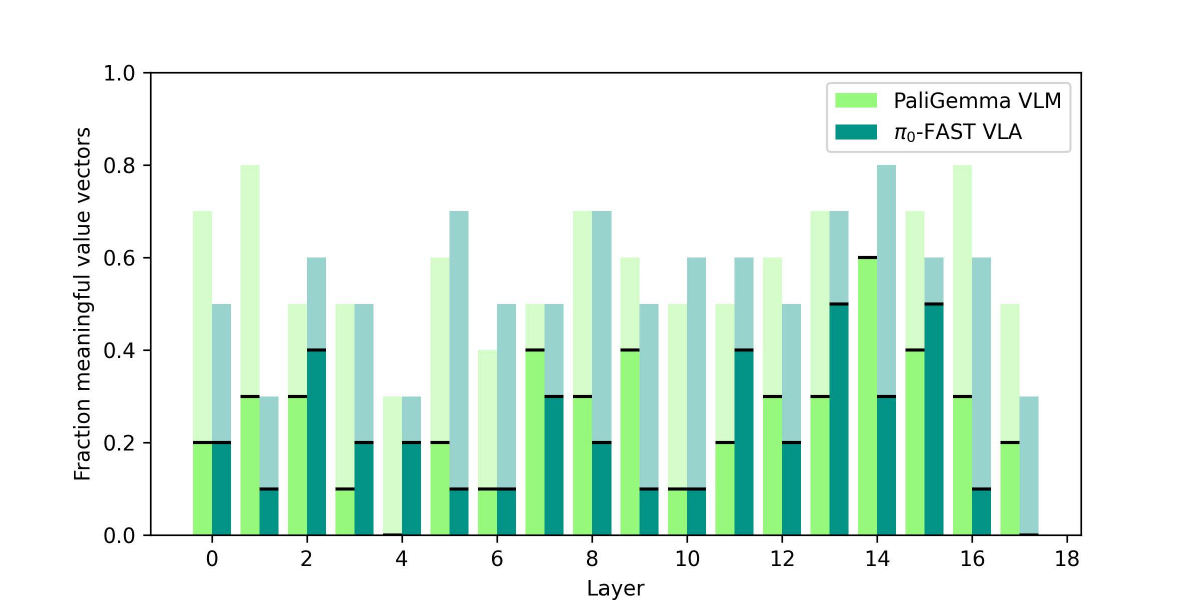}
        \caption{\textbf{Meaningful patterns in top value vector tokens}. VLA training does not substantially change the proportion of FFN value vectors which have interpretable patterns (top lighter bars) and semantically meaningful patterns (bold bottom bars) in their top tokens.}
        \label{fig:VLM_to_VLA_semantic}
    \end{subfigure}
    \hfill
    \begin{subfigure}[t]{0.36\textwidth}
        \centering
        \includegraphics[width=\linewidth]{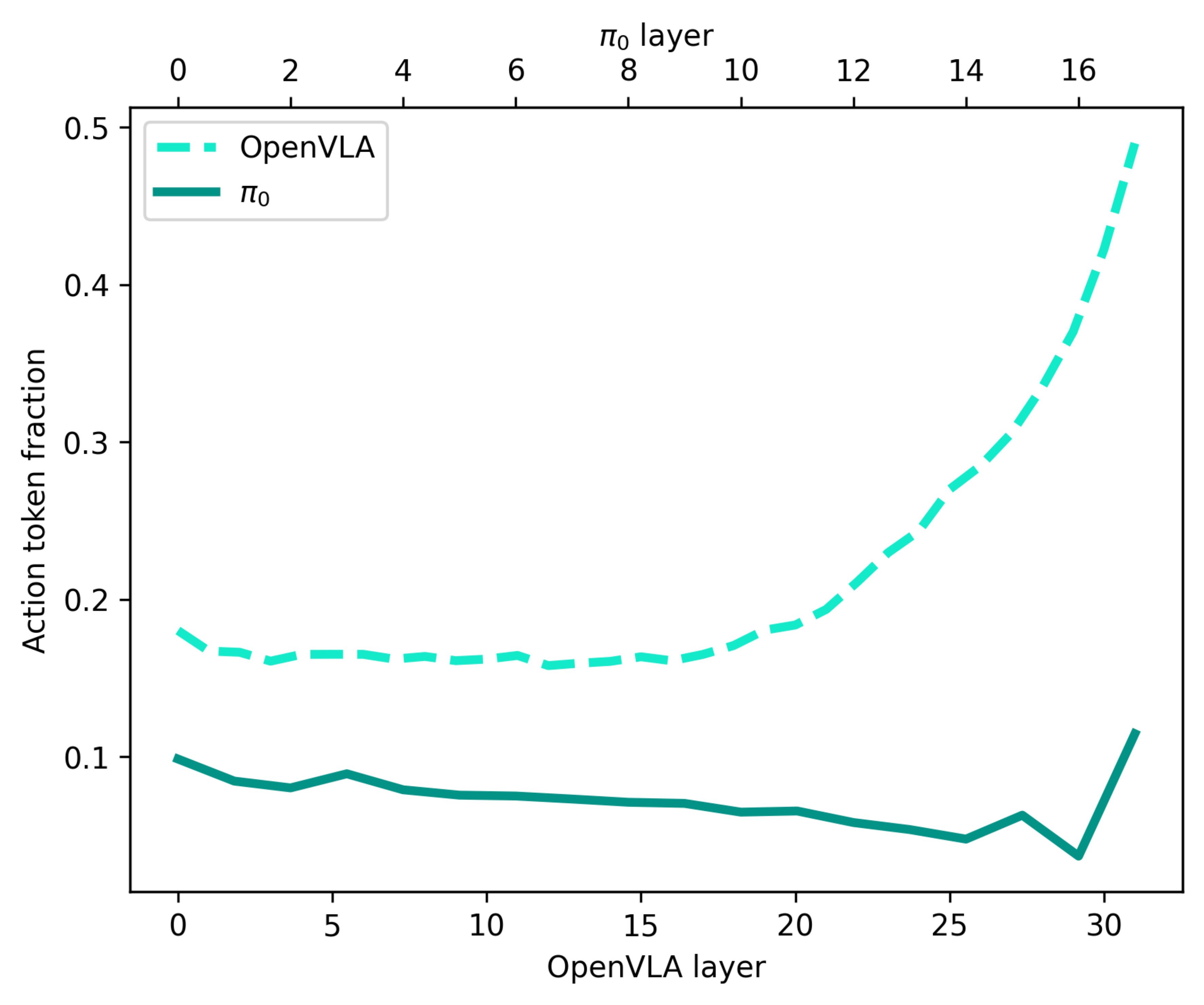}
        \caption{\textbf{Action tokens are incorporated into every layer of the VLA.} However, they make up the largest proportion of final layer value vectors.}
        \label{fig:VLA_actions_by_layer}
    \end{subfigure}
    \caption{VLA training incorporates action tokens into FFN value vectors but retains semantic meanings from VLM pre-training.}
    \vspace{-15pt}
\end{figure}

During VLA training, a pre-trained VLM is fine-tuned to imitate the control actions of expert robot trajectories conditioned on a task description and image and/or state observation, using a dataset such as  Open X-Embodiment \cite{embodimentcollaboration2024openxembodimentroboticlearning}. The most common strategy for handling the actions, followed by VLAs such as RT-2 \cite{brohan2023rt}, \textsc{OpenVLA} \cite{kimopenvla}, and $\pi_0$-FAST \cite{pertsch2025fast}, is to denote a small set of rarely used VLM tokens as ``action tokens'' to be decoded into control outputs. The final output of the model during VLA training is exclusively action tokens, which raises the question: does the VLA still use natural language-like representations to choose appropriate control actions? If so, how do language and action interact in the trained model? Our chosen interpretability technique allows us some insight into these questions:

\textbf{VLA Value Vectors Retain Semantic Meanings.} First, we aimed to quantify whether the VLM value vectors represent semantically interpretable concepts, and if so, whether this is preserved after VLA training. We focused on $\pi_0$ and its base pretrained VLM, PaliGemma-3B \cite{beyer2024paligemma}, because its larger vocabulary size, compared to \textsc{OpenVLA's} Llama2-7B base VLM \cite{touvron2023llama2openfoundation}, makes individual tokens easier to interpret. We randomly selected 10 value vectors from each layer of the VLM and VLA models and examined whether at least 4 of the top 30 tokens followed a common pattern (adopting the methodology from \cite{geva2022transformer}. The results are shown in Figure \ref{fig:VLM_to_VLA_semantic}. Value vectors throughout both the VLM and VLA models exhibit identifiable patterns, many of them semantic, at similar rates. Therefore, VLA training does not cause FFN outputs to lose meaningful organization of concepts, despite producing only non-semantic tokens at the final output. We show qualitative examples of value vectors in the Appendix.

\textbf{Action Tokens Appear in All VLA Layers.} We show the proportion of top-100 tokens which are action tokens for value vectors in each layer of the model (Figure \ref{fig:VLA_actions_by_layer}). For both \textsc{OpenVLA} and $\pi_0$, action tokens are most common at the final layer, but make up at least a few percent of every layer. This suggests that there is not a hard transition from ``thinking about the task'' in earlier layers to ``thinking about control'' in later layers. Instead, it appears VLA models learn to reason about and refine their control action predictions continuously, starting from their earliest computations.

\vspace{-2pt}
\subsection{How Task-Specific Fine-Tuning Affects VLA Concepts}
\begin{figure}[t]
    \centering
    \begin{subfigure}[t]{0.59\textwidth}
        \centering
        \includegraphics[width=\linewidth]{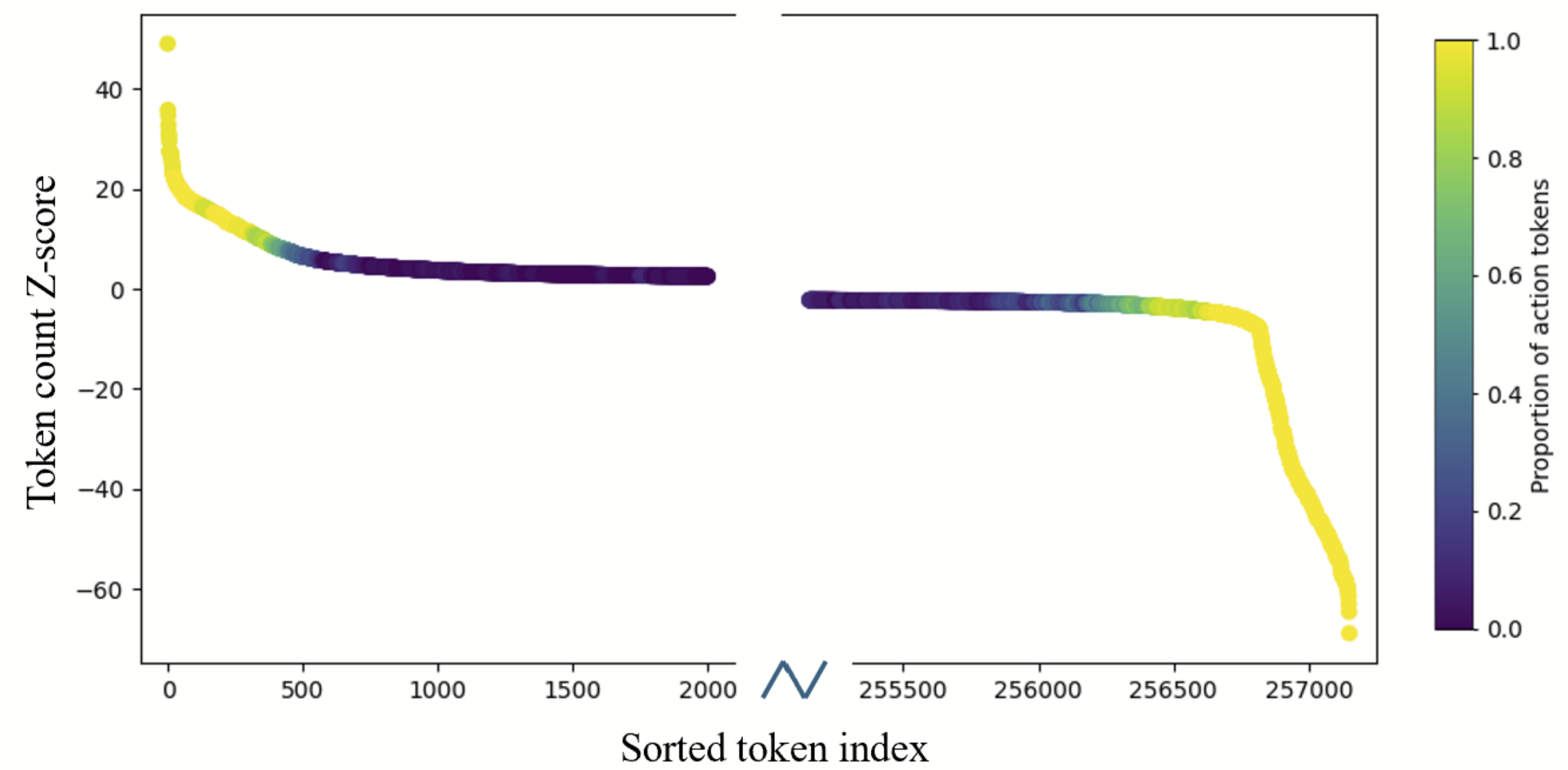}
        \caption{\textbf{Task fine-tuning mainly affects action tokens}. The most up-weighted and down-weighted tokens between the $\pi_0$-FAST and $\pi_0$-FAST-DROID-finetune models are action tokens.}
        \label{fig:droid_upweighted_tokens}
    \end{subfigure}
    \hfill
    \begin{subfigure}[t]{0.39\textwidth}
        \centering
        \includegraphics[width=\linewidth]{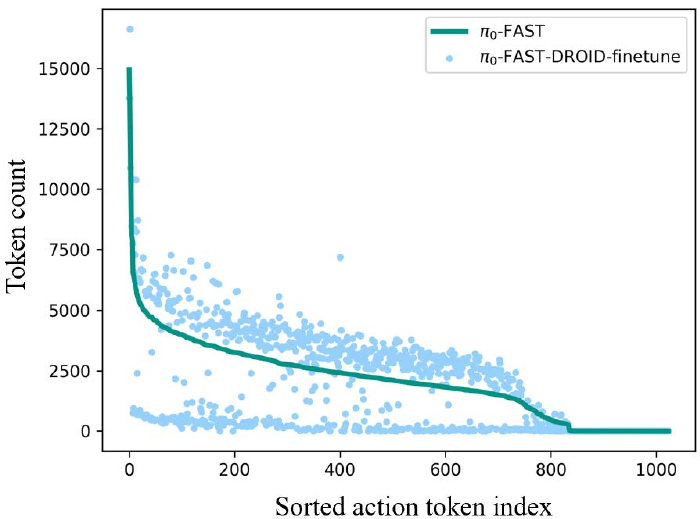}
        \caption{\textbf{Fine-tuning induces a more specialized (less general) distribution of action tokens across value vectors.}}
        \label{fig:droid_action_tokens}
    \end{subfigure}
    \caption{Task fine-tuning mainly affects action tokens in FFN value vectors.}
    \vspace{-15pt}
\end{figure}

\vspace{-1pt}
After VLA pre-training, models generally need to be fine-tuned for a specific robot setup and set of tasks to realize good performance. To understand what happens to the value vectors during fine-tuning, we compared the base $\pi_0$-FAST model with a checkpoint fine-tuned on the DROID dataset \cite{khazatsky2024droid}. We compared top tokens in the value vectors using a two-proportion z-test statistic to quantify the significance of an increase (or decrease) in the number of occurrences of each token between the two models. Our results are shown in Figure \ref{fig:droid_upweighted_tokens} and \ref{fig:droid_action_tokens}. We find that action tokens account almost exclusively for the most significant differences between the two sets of value vectors, and that this shift occurs because the fine-tuned model develops a more uneven, specialized distribution across action tokens compared to the relatively flat distribution of the pre-trained model. There is also a modest boost to some of the most common tokens in the DROID task instructions: these tokens appear 1.2 times more frequently in the fine-tuned model value vectors than in the pre-trained ones (see the Appendix for more details). Overall, these results suggest that the main effect of fine-tuning is to re-wire the model to more easily produce action tokens utilized in the fine-tuning tasks and neglect the others, with only a modest effect on semantic tokens.

\subsection{Summary}
The above results suggest that VLAs reason about control actions by mixing semantic concepts from pre-training with action tokens at all layers. However, these results do not prove a direct causal link between specific semantic concepts and robot behavior. If such a link exists, we expect to be able to \emph{intervene} on semantic concepts to change behavior. In the following section, we examine whether the semantic concepts we uncovered can support such interventions.

\section{Steering VLAs}
\label{sec:steering}

We formalize our method for \textit{interpretable activation-level steering} of VLAs (Figure~\ref{fig:main_graphic}). This technique modifies selected neurons within the FFN submodules of transformer blocks inside the VLA’s base VLM, steering robot behavior in real time - without fine-tuning or environment reward signal.

Let $x \in \mathbb{R}^{n}$ be the residual input to a transformer FFN. As shown in Equation~(\ref{eqn:FFN_decomposition}), the FFN output is a sum of fixed value vectors $w_\theta^{(i)} \in \mathbb{R}^n$ weighted by input-dependent activations $f_\theta(x) \in \mathbb{R}^m$. Motivated by evidence that individual FFN neurons encode semantic features \cite{geva2021transformer, bricken2023monosemanticity}, we override a subset $\mathcal{S} \subseteq \{1, \dots, m\}$ of activations using a fixed scalar $\alpha \in \mathbb{R}$. Each $\mathcal{S}$ corresponds to an interpretable neuron cluster aligned with a control concept - e.g., \textit{fast}, \textit{up}, \textit{careful} - identified by grouping neurons with similar token projections (Figure~\ref{fig:main_graphic}, steps 1–3), either manually or via kNN over semantic embeddings.
 
\vspace{-4mm}
\begin{equation}
\label{eqn:activation_override}
\tilde{f}_\theta^{(i)}(x) =
\begin{cases}
\alpha & \text{if } i \in \mathcal{S} \\
[f_\theta(x)]_i & \text{otherwise}
\end{cases}
\end{equation}
\vspace{0.2mm}
\begin{equation}
\label{eqn:steered_ffn}
\text{FFN}_{\text{steered}}(x) = \sum_{i=1}^{m} \tilde{f}_\theta^{(i)}(x) \cdot w_\theta^{(i)}
\end{equation}
\vspace{-3mm}

This induces a residual shift $\Delta x = \text{FFN}_{\text{steered}}(x) - \text{FFN}(x)$ that propagates through the transformer and modulates the final VLA action token distribution.

\noindent\textbf{Implementation.}
In \textsc{OpenVLA} (PyTorch), we apply a forward hook on the FFN's \texttt{down\_proj} to overwrite neuron activations. In $\pi_0$ (JAX), the neuron indices $\mathcal{S}$ and  activation coefficient $\alpha$ are passed into a refactored FFN. Both realize the same operator $\mathcal{I}_\mathcal{S}^\alpha(f_\theta(x))$ for real-time robot control.

\subsection{Simulation Experiments}
\label{sec:simulation_experiments}

\begin{figure*}
    \centering
    \includegraphics[width=\textwidth]{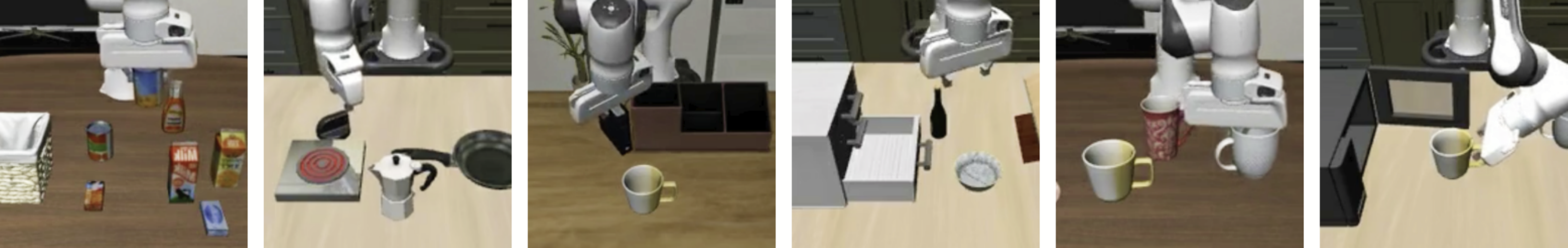}
    \caption{\textbf{Sample tasks from LIBERO-Long.} Six representative long-horizon tasks - involving sequential manipulation goals such as object placement, containment, and appliance interaction.}
    \label{fig:libero}
\end{figure*}

We evaluate our VLA steering method within the suite of ten long-horizon manipulation tasks making up the LIBERO-Long benchmark \cite{liu2023liberobenchmarkingknowledgetransfer}, visualized in Figure~\ref{fig:libero}. Rollouts are seeded deterministically for reproducibility, enabling precise comparisons between baseline and intervention trajectories. All simulation experiments use the 7B-parameter \textsc{OpenVLA} model, with the fine-tuned LIBERO-Long checkpoint, implemented in PyTorch and executed on a NVIDIA H100 GPU.

\begin{figure}[t]
    \centering
    \begin{subfigure}[t]{0.49\textwidth}
        \centering
        \includegraphics[width=\linewidth]{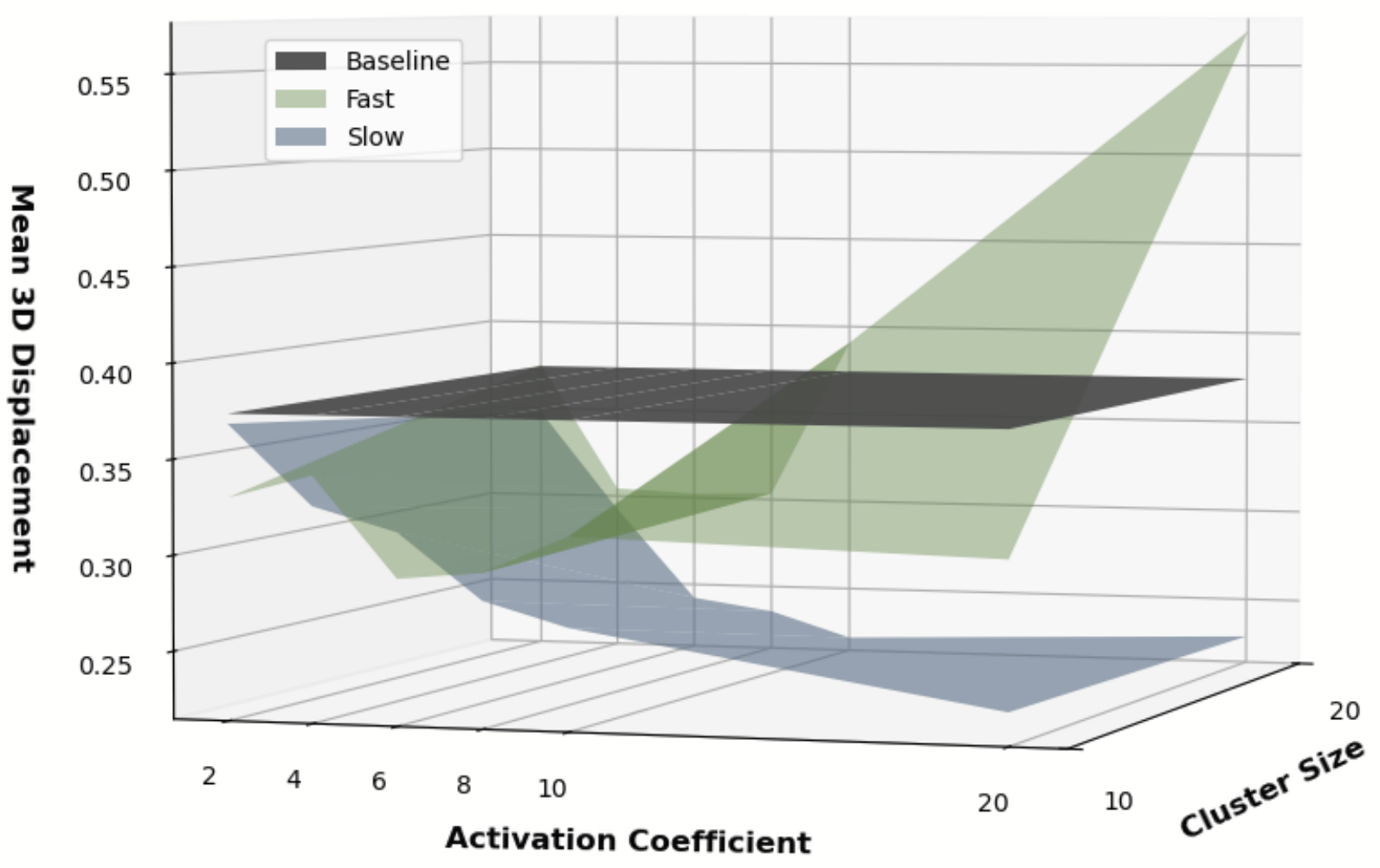}
        \caption{\textbf{Steering motion magnitude interventions.} The effect of fast and slow cluster interventions across cluster sizes and activation coefficients. Fast clusters consistently lead to larger end-effector displacement.}
        \label{fig:fastslow_surface}
    \end{subfigure}
    \hfill
    \begin{subfigure}[t]{0.48\textwidth}
        \centering
        \includegraphics[width=\linewidth]{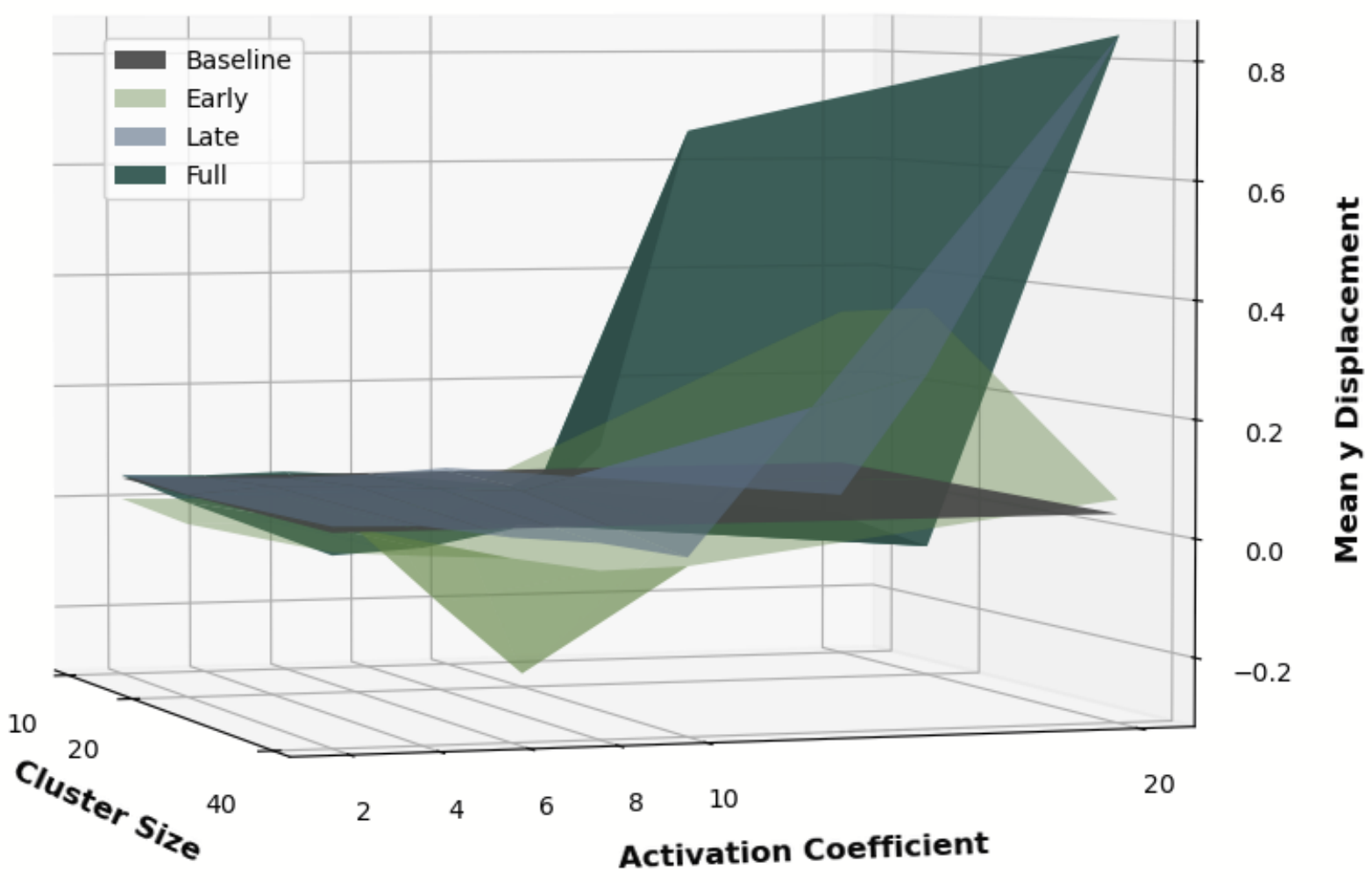}
        \caption{\textbf{Temporal localization interventions.} Mean Y-displacement for \textit{up}-cluster activations injected at early, late, and full model depths. Full clusters produce the largest average motion effects.}
        \label{fig:up_surface}
    \end{subfigure}
    \caption{\textbf{Simulation results:} Steering \textsc{OpenVLA} with value vector interventions.}
    \vspace{-15pt}
\end{figure}

\textbf{Steering Motion Interventions.} To evaluate whether interpretable semantic structure in VLA value vectors can steer physical behavior, we construct \textit{fast} and \textit{slow} aligned clusters by manually selecting value vectors whose top-projected tokens reflect motion magnitude semantics. These clusters are injected into the model’s residual stream with a fixed coefficient, sweeping over cluster sizes \([10, 20]\) and activation strengths \([2, 4, 6, 8, 10, 20]\). Each configuration is evaluated over 10 long-horizon tasks, using 10 rollouts per task. The fast clusters consistently induce larger end-effector displacements across tasks, with an average improvement of 27.73\% over slow clusters and maximum gains of 148.54\% in some configurations. All 10 matched comparisons yield statistically significant differences (\(p < 0.001\), paired t-test), and effect sizes (\(d = -0.091\) – \(1.419\)) suggest consistent directional influence. These results demonstrate that directional behavior can be modulated by activating semantically interpretable value vectors alone (Figure~\ref{fig:fastslow_surface}).

\textbf{Temporal Localization Interventions.} We examine how the position of activated value vectors within the VLA affects steering performance. Using kNN clustering over token-projected value vector embeddings, we extract \textit{up}-themed clusters and restrict their injection to early, late, or all layers. Each configuration spans three cluster sizes and six activation coefficients, with all other variables held constant. Across conditions, full-layer interventions yielded the largest average Y-displacements of the end-effector (\(\mu = 0.098\)), followed by late-layer (\(\mu = 0.086\)) and early-layer (\(\mu = 0.007\)) injections. Effect-size contrasts were most pronounced between early and late depths (\(d = -0.376\)), smaller between early and full (\(d = -0.321\)), and negligible between late and full (\(d = -0.062\)). As shown in Figure~\ref{fig:up_surface}, full-layer activations have the strongest effect on average, yet late-layer interventions can match them at higher intensities and larger cluster sizes. This is consistent with the expectation of explicit motion semantics concentrating in the model’s final stages.

\subsection{Robot Experiments}
\label{sec:robot_experiments}

We also evaluate our VLA steering method using the 3B-parameter $\pi_0$-FAST VLA on a real robot. Specifically, we test the ability to steer binary control opposites - \textit{slow vs.\ fast} and \textit{low vs.\ high} - in two pick-and-place tasks with a UR5 robot arm. We use a JAX implementation of $\pi_0$-FAST running on a NVIDIA A4500 GPU. Details on our custom robot setup are provided in the Appendix.

\textbf{Fine-Tuning.}  Unlike benchmarks such as DROID, which use a standardized robot platform and include fine-tuned $\pi_0$-FAST checkpoints, our robot setup is not represented in the $\pi_0$-FAST training data and cannot reasonably perform tasks zero-shot. To enable meaningful evaluation, we fine-tune with LoRA \cite{hu2021loralowrankadaptationlarge} on small datasets collected from our robot. Although fine-tuning on a limited dataset may hinder generalization and steering capability, it enables controlled experimentation: we manage the full data collection process and can introduce variations and behaviors in the robot data that we wish to steer during evaluation. Crucially, we do not explicitly label these variations in the VLA prompt during training, allowing us to test whether the VLA can reproduce these specific behaviors through steering by associating them with high-level semantic concepts. 

\begin{figure}[t]
    \centering
    \begin{subfigure}[t]{0.48\textwidth}
        \centering
        \includegraphics[width=\linewidth]{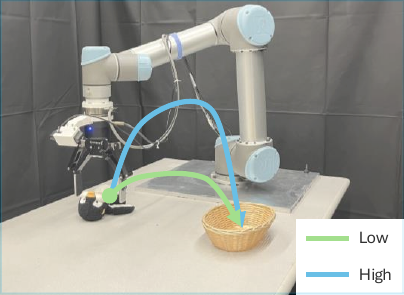}
        \caption{\textbf{Low/high transport.} The robot picks up a toy penguin and places it into a basket, with variations in the robot’s trajectory height during data collection.}
        \label{fig:low_high_transport_task}
    \end{subfigure}
    \hfill
    \begin{subfigure}[t]{0.48\textwidth}
        \centering
        \includegraphics[width=\linewidth]{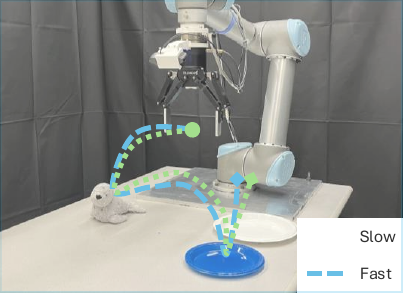}
        \caption{\textbf{Slow/fast transport.} The robot picks up a toy seal and places it onto a plate, with variations in the robot’s movement speed during data collection.}
        \label{fig:slow_fast_transport_task}
    \end{subfigure}
    \caption{\textbf{Physical robot experiments:} Steering $\pi_0$ on a UR5.}
    \vspace{-15pt}
\end{figure}

\textbf{Robot Tasks.}  
We evaluate steering on two tasks. The first task, \textit{Low/High Transport}, involves the robot picking up a toy penguin and placing it in a basket and includes 75 episodes with unlabeled variations in how high the penguin is lifted during transport. The second task, \textit{Slow/Fast Transport}, involves the robot picking up a toy seal and placing it on a specified plate and includes 120 episodes with variations in the speed of the robot's motion. Visualizations of both tasks are shown in Figures~\ref{fig:low_high_transport_task} and~\ref{fig:slow_fast_transport_task}, and data collection and fine-tuning details are provided in the Appendix.

For both tasks, we evaluate steering interventions where we hand-select and upweight a cluster of semantically meaningful vectors. We create a \textit{low} and \textit{high} themed cluster for \textit{Low/High Transport} and \textit{slow} and \textit{fast} themed cluster for \textit{Slow/Fast Transport}. We also compare these interventions against several baselines: (1) running the model without any intervention, (2) adding qualitative descriptors to the prompt (e.g. appending \textit{low} to the beginning of the prompt), and (3) applying steering with randomly upweighted vectors. These comparisons evaluate the model's default behavior without intervention, whether prompt modification alone can influence behavior, and whether selecting steering vectors based on semantic concepts are more effective than simply selecting random vectors. Specific details on each steering intervention and baseline is in the Appendix. 

\begin{figure}[t]
    \centering
    \begin{subfigure}[t]{0.49\textwidth}
        \centering
        \includegraphics[width=\textwidth]{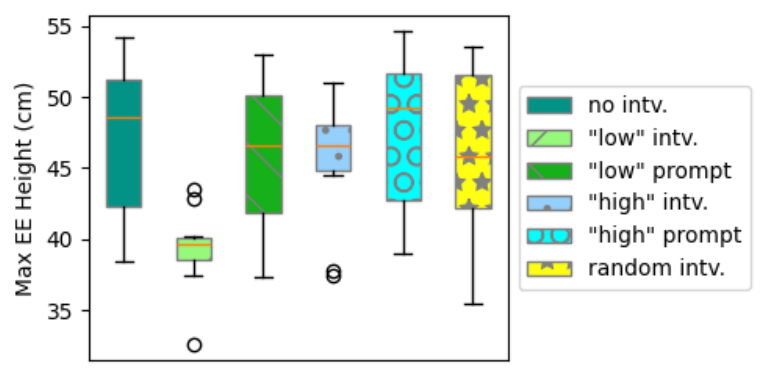}
        \caption{\textbf{Low/high transport.} Box plots for maximum end-effector height (cm) showing distribution across 10 rollouts for each steering intervention (intv.) and baseline.}
        \label{fig:low_high_transport_plot}
    \end{subfigure}
    \hfill
    \begin{subfigure}[t]{0.49\textwidth}
        \centering
        \includegraphics[width=\textwidth]{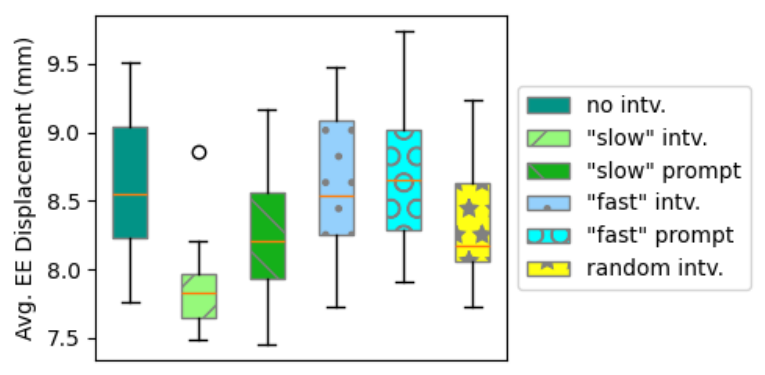}
        \caption{\textbf{Slow/fast transport.} Box plots for average end-effector displacement (mm) between each successive action showing distribution across 10 rollouts for each steering intervention (intv.) and baseline.}
        \label{fig:slow_fast_transport_plot}
    \end{subfigure}
    \caption{\textbf{Physical robot experiments:} Binary control opposites.}
    \vspace{-15pt}
\end{figure}

\textbf{Task Evaluation.}  
For \textit{Low/High Transport}, we perform 10 independent rollouts for each steering variant and baseline and analyze the maximum height of the end-effector relative to the robot’s base over the course of the task. We then record the maximum height across 10 rollouts for each variant in Figure~\ref{fig:low_high_transport_plot}. In \textit{Slow/Fast Transport}, we focus on measuring the robot's speed. To do this in a consistent manner we run inference for all steering variants and baselines simultaneously, but only execute the baseline predicted action with no intervention. We measure average end-effector displacement between each step along the trajectory across 10 rollouts for each variant and plot results in Figure~\ref{fig:slow_fast_transport_plot}. More detailed analysis of individual trajectories for both tasks are in the Appendix.

\textbf{Robot Results.}  
Our results suggest that steering interventions can meaningfully influence the actions of $\pi_0$-FAST. In the \textit{Low/High Transport} task, the \textit{low} intervention generated the lowest overall trajectories. Similarly, the \textit{slow} intervention in \textit{Slow/Fast Transport} resulted in the slowest overall movements. However, while \textit{low} and \textit{slow} interventions clearly altered behavior, their opposites (\textit{high} and \textit{fast}) resembled the baseline without intervention. This may be because the model already ``considers'' the baseline trajectory to be fast and high. Nevertheless, the minimal difference between using a random intervention compared to no intervention suggests that choosing semantically meaningful vectors can more effectively guide the model towards desired behavior. Finally, we found that steering interventions were more impactful than simply adding target descriptive words to the prompt. In both tasks, modifying the prompt had weaker impact than our steering intervention. 

\section{Discussion and Conclusion}
\label{sec:conclusion}

\textbf{This work shows that VLAs can be steered via interpretable internal components.} By activating sparse, semantically meaningful neurons in FFN layers---aligned with concepts like ``up'' and ``slow''---we modulate robot behavior at inference time. Experiments demonstrate that our method generalizes across models and tasks, influencing both simulated and real-world behavior, including long-horizon LIBERO tasks and binary control opposites on a UR5.

\textbf{From a mechanistic perspective}, this work provides causal evidence that VLAs retain structured, compositional semantics even after fine-tuning. That activating a small set of semantically aligned neurons can shift motion behavior suggests control-relevant features remain accessible and manipulable. These effects emerge without explicit behavioral supervision, indicating that VLAs internalize semantic concepts during pretraining in a way that supports compositional reasoning across layers.

\textbf{For robotics practitioners}, this method offers a simple, interpretable tool for post-deployment behavior shaping. It requires no retraining, runs on existing policies, and enables fast testing and adjustment. Our experiments show that it can modulate some aspects of robot behavior---such as end-effector speed or height---and suggest that this modulation may be more effective than using modified prompts.

\textbf{Looking ahead}, activation steering opens a new interpretable control interface for guiding VLA behavior. As embodied foundation models become more capable, steering techniques could play an important role in inspecting, auditing, and intervening in how models reason and act. We see this approach---using just one tool from the mechanistic interpretability toolbox---as a step toward a broader class of transparent, steerable, and semantically grounded control tools for robotics.


\section{Limitations}
\label{sec:limitations}

This work introduces one of the first interpretable control knobs for foundation models in robotics, enabling behavior to be modulated through semantically grounded activations without retraining. Establishing such control-layer transparency is a step toward safe, adaptive embodied AI. While the results are promising, several challenges remain, including reliably mapping internal structure to behavior and scaling to more general settings. Addressing these challenges presents concrete opportunities to advance the robustness, generality, and practical utility of activation-based steering.

\textbf{Semantic Ambiguity and Representational Drift.} Our clustering method is based on token-level semantic similarity rather than direct behavioral outcomes. For example, approaches like kNN over token-projected value vectors assume that local similarity implies shared semantics. In practice, clusters can conflate distinct behaviors---e.g., ``slow and careful'' versus ``slow and stuck''---leading to inconsistent effects. Furthermore, we found that the meaning of value vectors can shift across models, tasks, and time. A vector aligned with “up” in one context may behave differently in another. Understanding and mitigating this representational drift will be critical for developing steering interventions that transfer reliably across domains and deployment scenarios.

\textbf{Fine-Tuning and Steerability.} We do not yet fully understand how VLA fine-tuning affects steerability. While some semantic directions appear to retain causal influence after adaptation, it's possible that fine-tuning alters internal representations in ways that make VLM-pretrained concepts harder to access or less aligned with behavior. Investigating how steerable directions evolve with continued training is an ideal direction for future work.

\textbf{Evaluation Scope and Generalization.} Our evaluations, while spanning both simulation and hardware, are limited to pick-and-place tasks with a robot arm. Extending to mobile and bimanual platforms, as well as unstructured environments, will be important for validating control stability, generalization under physical variation, and alignment with human intent across real-world settings.

\clearpage

\acknowledgments{

This work is supported by the NSF Safe Learning Enabled Systems Program, the DARPA Assured Neuro Symbolic Learning and Reasoning Program, and the ONR project ``Leveraging Egocentric and Allocentric Representations for Navigation (LEARN)''.

Bear H\"{a}on was additionally supported by the Schmidt Futures Quad Fellowship, NSF DToD Fellowship, and Foresight Institute Fellowship during the development of this work. He thanks the ERA Fellowship at the University of Cambridge for early exposure to mechanistic interpretability during his time as a 2023 AI Safety Fellow.\\

}


\bibliography{example}  

\clearpage

	
\appendix

\section{Steering Intervention Details}

Let $x \in \mathbb{R}^{n}$ be the residual input to a transformer FFN. As shown in Equation~(\ref{eqn:FFN_decomposition}), the FFN output is a sum of fixed value vectors $w_\theta^{(i)} \in \mathbb{R}^n$ weighted by input-dependent activations $f_\theta(x) \in \mathbb{R}^m$. Motivated by evidence that individual FFN neurons encode semantic features \cite{geva2021transformer, bricken2023monosemanticity}, we override a subset $\mathcal{S} \subseteq \{1, \dots, m\}$ of activations using a fixed scalar $\alpha \in \mathbb{R}$. Each $\mathcal{S}$ corresponds to an interpretable neuron cluster aligned with a control concept - e.g., \textit{fast}, \textit{up}, \textit{careful} - identified by grouping neurons with similar token projections (Figure~\ref{fig:main_graphic}, steps 1–3), either manually or via kNN over semantic embeddings.

\vspace{-4mm}
\begin{equation}
\label{eqn:activation_override}
\tilde{f}_\theta^{(i)}(x) =
\begin{cases}
\alpha & \text{if } i \in \mathcal{S} \\
[f_\theta(x)]_i & \text{otherwise}
\end{cases}
\end{equation}
\vspace{0.2mm}
\begin{equation}
\label{eqn:steered_ffn}
\text{FFN}_{\text{steered}}(x) = \sum_{i=1}^{m} \tilde{f}_\theta^{(i)}(x) \cdot w_\theta^{(i)}
\end{equation}
\vspace{-3mm}

This induces a residual shift $\Delta x = \text{FFN}_{\text{steered}}(x) - \text{FFN}(x)$ that propagates through the transformer and modulates the final VLA action token distribution.

\noindent\textbf{Implementation.}
In both $\pi_0$ and \textsc{OpenVLA}, the FFN layers use a GEGLU activation function \cite{shazeer2020glu} which has the form $[f_\theta(x)]_i = [\text{GELU}(W_1 x)]_i \cdot [W_2 x]_i$, where $W_1$, $W_2 \in \mathbb{R}^{m x n}$ are additional parameter matrices. To implement the steering intervention, for $\pi_0$ (JAX), the neuron indices $\mathcal{S}$ and activation coefficient $\alpha$ are passed into modified FFN code that implements Equation~(\ref{eqn:activation_override}). For \textsc{OpenVLA} (PyTorch), we apply a forward hook on the FFN's \texttt{down\_proj} to overwrite activations. 

We also experimented with a modified intervention to \textsc{OpenVLA} which instead applies the forward hook to the FFN's \texttt{gate\_proj}. This is equivalent to the following:

\begin{equation}
    \tilde{f}_\theta^{(i)}(x) =
    \begin{cases}
    \text{GELU}(\alpha) \cdot [W_2 x]_i & \text{if } i \in \mathcal{S} \\
    [f_\theta(x)]_i & \text{otherwise}
    \end{cases}
\end{equation}

We found that this intervention performed similarly to the intervention defined in Equation~(\ref{eqn:activation_override}) during our exploratory experiments. 

\section{Interpretability Experiments}
\subsection{Patterns in Value Vectors}

Following \cite{geva2022transformer}, we classified a value vector as meaningful if 4 of its top 30 predicted tokens followed a common pattern. Furthermore, we classified patterns as ``non-semantic'' if they could be explained by a shallow syntax-level rule, such as words starting with the same 3 letters, and ``semantic'' otherwise. If multiple different patterns were found, we selected the one containing the highest-ranked token for determining the classification of the vector. Some examples of vectors from each category are show in table \ref{tab:paligemma_vectors} (PaliGemma VLM) and \ref{tab:pi0_vectors} ($\pi_0$ VLA). Both models have 18 layers.

\begin{table}[h]
\centering
\renewcommand{\arraystretch}{1.4}  
\small  

\begin{subtable}[h]{\textwidth}
\centering
\begin{tabularx}{\textwidth}{|c|c|p{2.8cm}|X|}
\hline
\textbf{Layer} & \textbf{Pattern type} & \textbf{Pattern description} & \textbf{Top 10 Tokens} \\
\hline
  3 & None & N/A & strick, campa, laun, increa, accla, reluct, embra, exem, desir, fath \\
  6 & Non-semantic & Camel case variable names &\textless bos\textgreater, EndGlobalSection, Clik, expandindo, SourceChecksum, NSCoder, EditorBrowsable, UnusedPrivate, nahilalakip, ScopeManager \\
  6 & Semantic & Image-sourcing websites & pixabay, gettyimages, shutterstock, unsplash, uefa, Ferdin, plak, makro, Hæ, Lombar \\
  14 & Semantic & Mice and keyboards & mouse, keyboard, keys, KEY, keyboard, Mouse, keys, Mouse, mouse, clavier \\
\hline
\end{tabularx}
\caption{Example value vectors and their top tokens from the PaliGemma VLM.}
\label{tab:paligemma_vectors}
\end{subtable}

\begin{subtable}[h]{\textwidth}
\centering
\begin{tabularx}{\textwidth}{|c|c|p{2.7cm}|X|}
\hline
\textbf{Layer} & \textbf{Pattern type} & \textbf{Pattern description} & \textbf{Top 10 Tokens} \\
\hline
  1 & Non-semantic & Action tokens & \texttt{\textless Ac0701\textgreater, \textless Ac0689\textgreater, \textless Ac0706\textgreater, \textless Ac0695\textgreater, \textless Ac0714\textgreater, \textless Ac0719\textgreater, \textless Ac0684\textgreater, \textless Ac0688\textgreater, \textless Ac0305\textgreater, \textless Ac0334\textgreater}  \\
  14 & None & N/A & maneu, shenan, affor, impra, increa, encomp, fortn, accla, philanth, volunte \\
  15 & Semantic & Strong & tough, strength,tough, toughness, toughest, tougher, strong,strength, strongest,strong \\
  18 & Semantic & Given names & javier, jorge, alberto, felipe, Áng, ricardo, eduardo, roberto, sergio, fernando \\
\hline
\end{tabularx}
\caption{Example value vectors and their top tokens from the $\pi_0$ VLA.}
\label{tab:pi0_vectors}
\end{subtable}

\end{table}

\subsection{DROID Fine-tuning}

\begin{figure}[ht]
    \centering
    \includegraphics[width=0.7\linewidth]{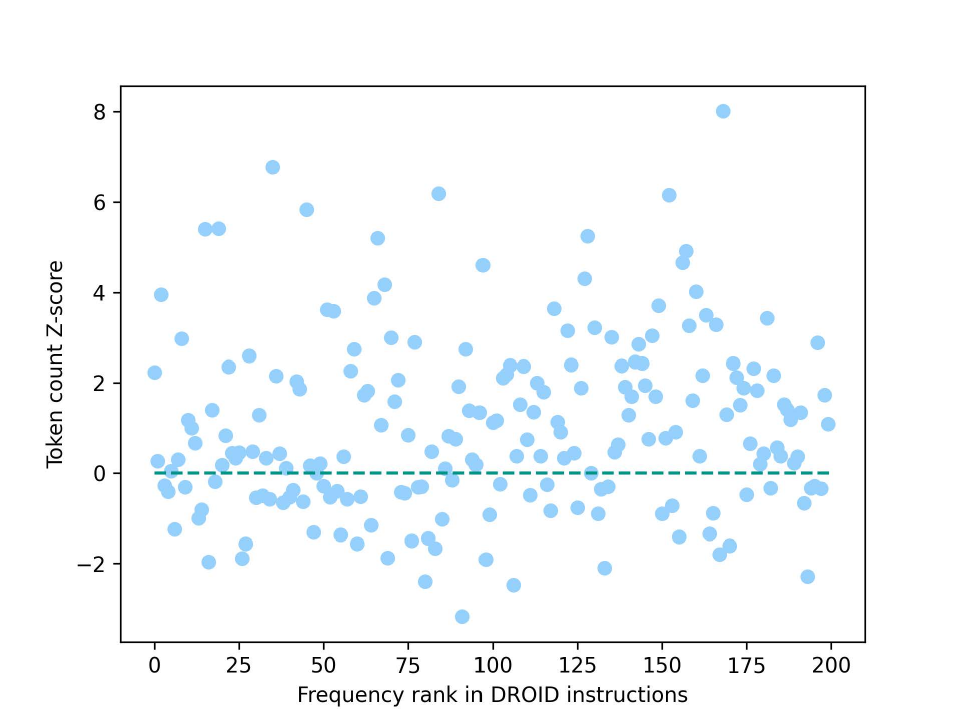}
    \caption{\textbf{Common DROID instruction tokens become more common in the value vectors of a model fine-tuned on the DROID dataset.} The majority of the top-200 instruction tokens are more common in the value vectors of the $\pi_0$-FAST-DROID-finetune checkpoint, corresponding to a positive z-score (lying above the teal dashed line).}
    \label{fig:droid_instruct_tokens}
    \vspace{-15pt}
\end{figure}

To understand whether the contents of the task instructions in the DROID training dataset affected the value vectors of the model checkpoint fine-tuned on DROID data, we examined whether common instruction tokens became more common in the value vectors after fine-tuning. Specifically, we first tokenized all task instructions (including multiple alternative instructions for the same demonstration where available) using the PaliGemma tokenizer. We then took the top 200 most frequent instruction tokens and examined the number of times they appeared in the top-100 tokens for each value vector in the $\pi_0$-FAST and $\pi_0$-FAST-DROID-finetune model checkpoints. We assign a z-score to each token based on a two-proportion z-test and plot the results in \ref{fig:droid_instruct_tokens}. A positive z-score indicates the token is more common in the value vectors of $\pi_0$-FAST-DROID-finetune compared to $\pi_0$-FAST, while a negative score indicates that it is more rare in $\pi_0$-FAST-DROID-finetune. We find that the majority of the top instruction tokens are indeed more common in the $\pi_0$-FAST-DROID-finetune value vectors, with a mean z-score of 1.0 and a mean count ratio of 1.2. While some tokens do have negative z-scores, we hypothesize that these tokens may be even more common in the base $\pi_0$ training data instructions, which unfortunately we do not have access to. Overall, these results suggest that semantic instruction inputs during VLA training may contribute to these tokens being retained in the value vectors, despite the VLA model being trained to only output action tokens.

\section*{Simulation Experiments}
\subsection{KNN for Temporal Depth Interventions}

\textbf{1. Semantic Embedding Construction.}  
To assign meaning to each value vector, we project it into the model’s output token space using the language modeling head, as described in Section~\ref{sec:interpreting}. For each value vector $w_\theta^{(i)}$, we compute token logits and identify the top-$5$ tokens it most strongly activates. We then construct a semantic embedding by computing the softmax-weighted average of those token embeddings:
\[
\mathbf{e}_{\text{sem}}^{(i)} = \sum_{j\in t_i} \text{softmax}(l_j^{(i)}) \cdot W_j
\]
where $t_i$ are the top token indices for value vector $w_\theta^{(i)}$, $l_j^{(i)}$ is the logit that $w_\theta^{(i)}$ assigns to token $j$, and $W_j$ is the output embedding vector corresponding to token $j$. This yields a normalized, interpretable embedding for each value vector.

\textbf{2. Depth Partitioning.}  
The full \textsc{OpenVLA} model contains 352,255 value vectors across all transformer layers. To analyze temporal structure, we partition them: the first half (176,128 vectors) represents \emph{early-layer} vectors, while the second half represents \emph{late-layer} vectors.

\textbf{3. kNN Clustering.}  
We apply cosine-based k-nearest neighbor (kNN) clustering to these semantic embeddings using the cuML GPU-accelerated library. Each value vector is assigned a cluster based on its $k$ nearest neighbors ($k \in \{10, 20, 40\}$). For each cluster, we compute a semantic centroid by averaging its members’ embeddings.

\textbf{4. Concept-Aligned Cluster Selection.}  
To find clusters aligned with a target concept (e.g., ``up''), we tokenize the concept word/phrase using the model’s tokenizer, embed it using the language modeling head, and compare it to all cluster centroids using cosine similarity. The cluster with the highest similarity is selected for activation.

\textbf{5. Temporal Partitioning for Layer-Specific Interventions.}  
To isolate temporal effects, the above clustering pipeline was run independently on the \emph{early-layer} and \emph{late-layer} vector subsets. This ensured that the selected clusters were sourced exclusively from the intended depth region. For full-model interventions, clustering was performed on the full set of value vectors.

\begin{figure}[h]
\centering
\begin{tabularx}{\textwidth}{|c|X|}
\hline
\textbf{Vector} & \textbf{Top 5 Tokens} \\
\hline
153784 & up, [action: 0.428], Up, oks, cker \\
263055 & up, Up, Up, up, pto \\
276423 & up, coming, Up, up, forth \\
225616 & up, left, Up, up, Up \\
274043 & up, Up, Up, up, UP \\
285310 & up, Up, Up, up, UP \\
287766 & up, Up, Up, up, UP \\
287383 & up, Up, Up, up, UP \\
263704 & Up, up, Up, up, Down \\
323343 & up, Up, Up, up, UP \\
\hline
\end{tabularx}
\caption{Top-5 token projections for the most semantically aligned \textbf{``up''} clusters from the full model using kNN clustering with $k=10$.}
\label{fig:up_clusters_full_k10}
\end{figure}

\section{Hardware Experiments}

\begin{figure}[h]
  \centering
  \includegraphics[width=1.0\textwidth]{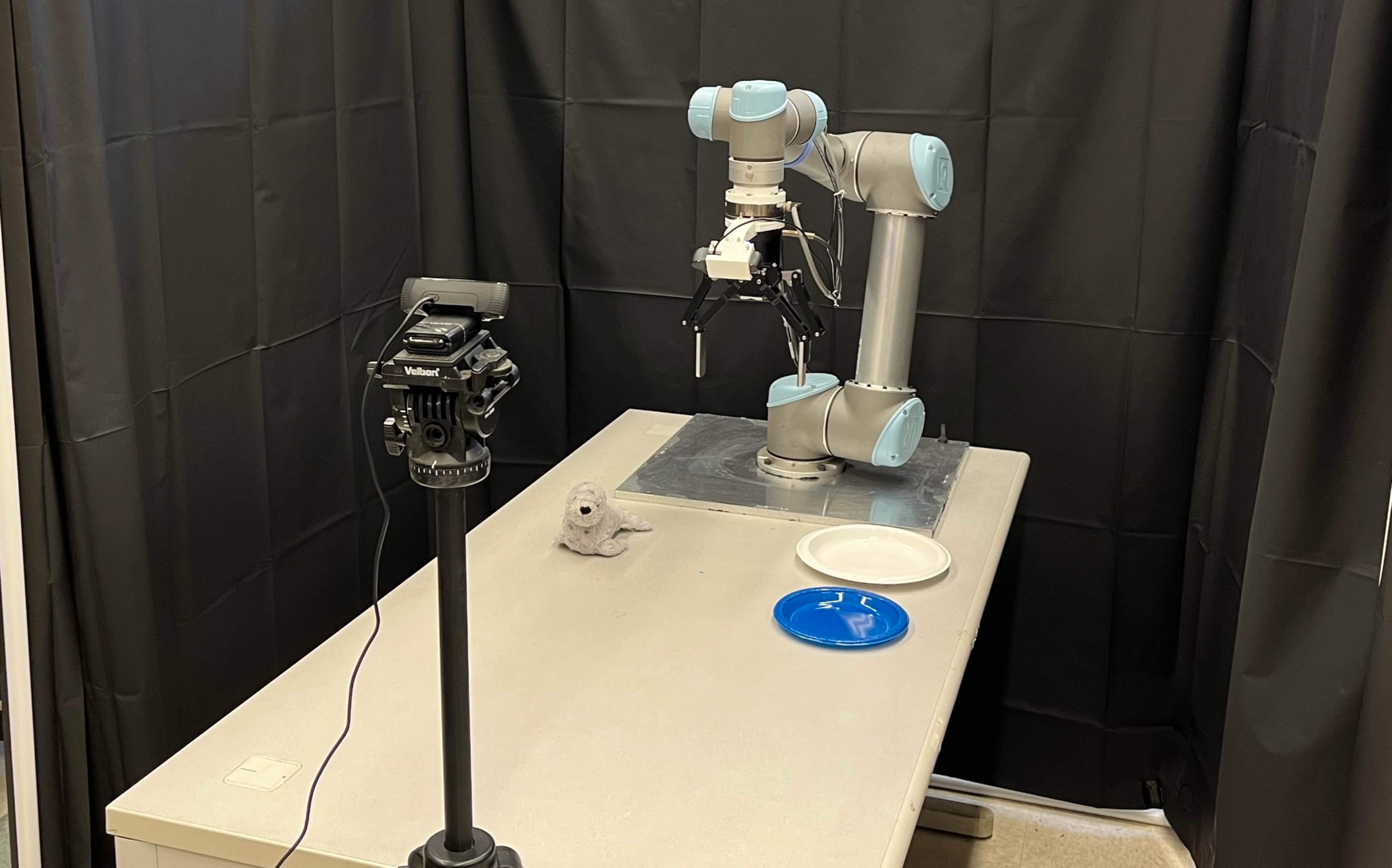}
    \caption{\textbf{Robot Setup:} Our hardware experiments use a UR5 robot arm equipped with a Robotiq 2F-140 gripper. The setup includes two cameras: a static scene camera overlooking the workspace and a wrist-mounted camera facing the gripper.}
  \label{fig:robot_setup}
\end{figure}

\subsection{Data Collection Details}

\begin{figure}[h]
    \centering
    \begin{subfigure}[t]{0.48\textwidth}
        \centering
        \includegraphics[width=\linewidth]{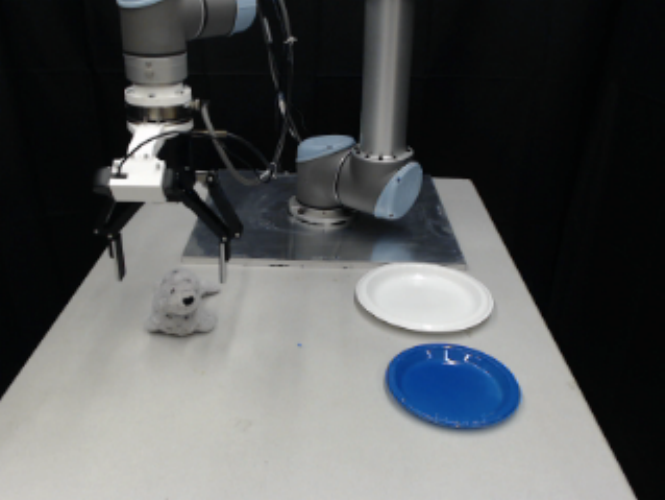}
        \caption{Scene Camera}
        \label{fig:robot_scene_camera}
    \end{subfigure}
    \hfill
    \begin{subfigure}[t]{0.48\textwidth}
        \centering
        \includegraphics[width=\linewidth]{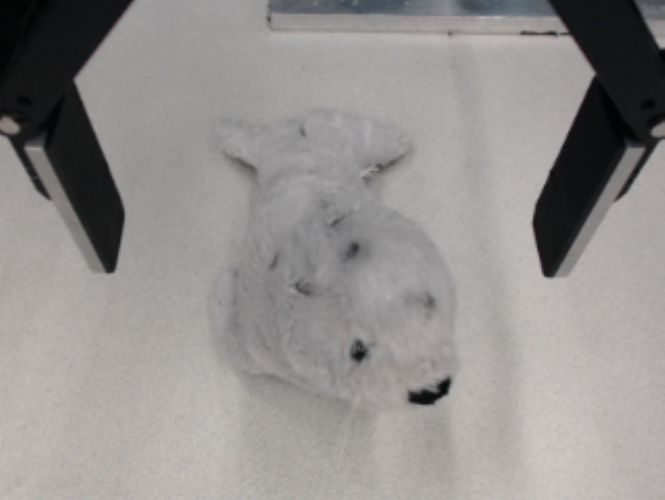}
        \caption{Wrist Camera}
        \label{fig:robot_wrist_camera}
    \end{subfigure}
    \caption{Example of raw images that are input into $\pi_0$-FAST.}
    \label{fig:robot_camera_images}
\end{figure}

During data collection and evaluation, objects for both tasks are placed within a consistent area with slight variations in position. In both tasks, the robot begins from the same initial pose with a small random deviation in position. During data collection, the end-effector is controlled in Cartesian space using a joystick. Data is recorded at 10 Hz, including images from the wrist-mounted and scene cameras, arm joint positions, gripper position, and the corresponding target joint and gripper commands at each timestep. Example images of both cameras are shown in Figure \ref{fig:robot_camera_images}.

For \textit{Low/High Transport}, we collect 75 episodes in total—25 each where the robot transports the toy penguin at a low, medium, or high height. Each episode is labeled with the prompt, ``place penguin in basket''. For \textit{Slow/Fast Transport}, we collect 20 episodes each with a slow, medium, or fast speed and placing the seal on either a blue or white plate, totaling 120 episodes. Each episode is labeled either with the prompt ``place seal on blue plate'' or ``place seal on white plate'' depending on which plate color the seal is placed on.

\subsection{Finetuning $\pi_0$-FAST}

We use LoRA for fine-tuning $\pi_0$-FAST and adopt the default training configuration provided by the official $\pi_0$-FAST codebase\textsuperscript{1}\footnote{\textsuperscript{1}\url{https://github.com/Physical-Intelligence/openpi}}. Fine-tuning is performed separately for each task, resulting in two distinct models—one for the \textit{Low/High Transport} task and one for the \textit{Slow/Fast Transport} task. For both tasks, the model input consists of the UR5’s six joint positions, the gripper position, and both scene and wrist camera images. The model outputs an action chunk as a sequence of joint and gripper positions. We fine-tune each model for 5000 steps using a batch size of 32 and an action chunk size of 10. We use preset normalization values provided by $\pi_0$-FAST specifically for the UR5e robot arm. 

\subsection{Steering Intervention and Baselines}

\textbf{Steering Intervention:} We hand-select and upweight a cluster of semantically meaningful value vectors by a coefficient of 10. Since each value vector can be interpreted as a probability distribution over output tokens, we examine the top-10 tokens with the highest probabilities for each vector. We then identify the top six value vectors containing the highest frequency of task-relevant keywords. For \textit{Low/High Transport} we search for the keyword ``low'' for the low intervention and ``high'' for the high intervention. For \textit{Slow/Fast Transport} we search for keywords ``slow'' and ``safe'' for the slow intervention and ``fast'' and ``risk'' for the fast intervention. We provide the value vectors used in the hardware experiments and their top-10 tokens in Table \ref{tab:combined_vectors}.

\textbf{Random Intervention Baseline:} We randomly select a cluster of six value vectors and upweight them by a coefficient of 10. We keep these vectors the same for the whole evaluation.

\textbf{No Intervention Baseline:} We run inference on $\pi_0$-FAST without any modification and use the same prompt as during fine-tuning: ``place penguin in basket'' for \textit{Low/High Transport} and ``place seal on blue plate'' for \textit{Slow/Fast Transport}.

\textbf{Prompt Modification Baseline:} We prepend steering-related keywords to the original prompts used during finetuning. For \textit{Low/High Transport}, the prompt “place penguin in basket” is modified to “low place penguin in basket” or “high place penguin in basket” for the low and high interventions, respectively. For \textit{Slow/Fast Transport}, the prompt “place seal on blue plate” is modified to “slow safe place seal on blue plate” or “fast risk place seal on blue plate” for the slow and fast interventions, respectively.

\begin{table}[h]
\centering
\renewcommand{\arraystretch}{1.4}  
\small  

\begin{subtable}[h]{\textwidth}
\centering
\begin{tabularx}{\textwidth}{|c|X|}
\hline
\textbf{Vector} & \textbf{Top 10 Tokens} \\
\hline
  5674 & low, low, Low, Low, lower, LOW, lower, lowest, LOW, lows \\
253282 & low, low, Low, Low, LOW, LOW, Lower, lower, Lower, lowest \\
263305 & low, high, low, high, High, High, LOW, Low, Low, LOW \\
230595 & low, medium, LOW, high, low, MEDIUM, Low, Low, Medium, medium \\
256528 & low, Low, fillType, low, Low, StoreMessageInfo, PerformLayout, LOW, lows, WithIOException \\
246672 & low, lows, low, Darío, cushi, Valentín, ecru, LOW, LOW, chaub \\
\hline
\end{tabularx}
\caption{Low intervention for \textit{Low/High Transport}.}
\label{tab:low_vectors}
\end{subtable}

\begin{subtable}[h]{\textwidth}
\centering
\begin{tabularx}{\textwidth}{|c|X|}
\hline
\textbf{Vector} & \textbf{Top 10 Tokens} \\
\hline
260217 & high, high, High, High, HIGH, HIGH, low, low, Low, LOW \\
265014 & high, high, High, High, HIGH, HIGH, hight, shenan, intersper, \begin{CJK*}{UTF8}{gbsn}高\end{CJK*} \\
266729 & high, high, High, High, HIGH, HIGH, highs, \begin{CJK*}{UTF8}{gbsn}高\end{CJK*}, middle, hight \\
267454 & high, High, High, high, HIGH, hight, HIGH, Odys, hig, quitted \\
270442 & high, high, High, High, HIGH, HIGH, hight, higher, inappro, unwarran \\
275165 & High, high, high, High, HIGH, HIGH, yüksek, \begin{CJK*}{UTF8}{gbsn}高\end{CJK*}, hoog, hight \\
\hline
\end{tabularx}
\caption{High intervention for \textit{Low/High Transport}.}
\label{tab:high_vectors}
\end{subtable}

\begin{subtable}[h]{\textwidth}
\centering
\begin{tabularx}{\textwidth}{|c|X|}
\hline
\textbf{Vector} & \textbf{Top 10 Tokens} \\
\hline
242252 & safe, safe, safely, Safe, Safe, SAFE, SAFE, safety, safer, \textless bos\textgreater \\
104785 & çük, SLOW, slow, Slow, slow, Slow, maksi, lacable, jät, tempo \\
243496 & safe, safe, SAFE, ArgumentParser, Safe, Safe, \textless bos\textgreater, Confira, memoized, Ainda \\
 73159 & soeur, slow, slowest, calm, Slow, slow, SLOW, calmer, hairc, atguigu \\
230490 & safe, safety, safe, Safe, Safe, Safety, Safety, safety, safely, SAFETY \\
234025 & slow, slowly, slow, \textless bos\textgreater, \textless Ac1403\textgreater, \textless Ac1327\textgreater, Slow, Slow, \textless Ac1340\textgreater, \textless Ac1450\textgreater \\
\hline
\end{tabularx}
\caption{Slow intervention for \textit{Slow/Fast Transport}.}
\label{tab:slow_vectors}
\end{subtable}

\begin{subtable}[h]{\textwidth}
\centering
\begin{tabularx}{\textwidth}{|c|X|}
\hline
\textbf{Vector} & \textbf{Top 10 Tokens} \\
\hline
239787 & risk, risk, \textless bos\textgreater, Risk, Risk, RISK, risks, Risks, RISK, riesgo \\
240184 & fast, fast, FAST, Fast, Fast, FAST, quickly, tolerably, hentai, Quickly \\
259344 & risk, risk, Risk, Risk, risks, Risks, riesgo, RISK, RISK, risky \\
262123 & fast, fast, Fast, Fast, faster, FAST, fastest, quick, quicker, FAST \\
271027 & fast, fast, fastest, faster, Fast, Fast, FAST, Faster, faster, FAST \\
273133 & fast, fast, Fast, Fast, FAST, FAST, \textless Ac1446\textgreater, faster, \textless Ac0390\textgreater, veloce \\
\hline
\end{tabularx}
\caption{Fast intervention for \textit{Slow/Fast Transport}.}
\label{tab:fast_vectors}
\end{subtable}

\caption{Value vectors and their top-10 highest probability output tokens for steering interventions in \textit{Low/High Transport} and \textit{Slow/Fast Transport}.}
\label{tab:combined_vectors}
\end{table}

\clearpage

\subsection{Steering Intervention Trajectories}

\begin{figure}[h]
    \centering
    \begin{subfigure}[t]{0.49\textwidth}
        \centering
        \includegraphics[width=\textwidth]{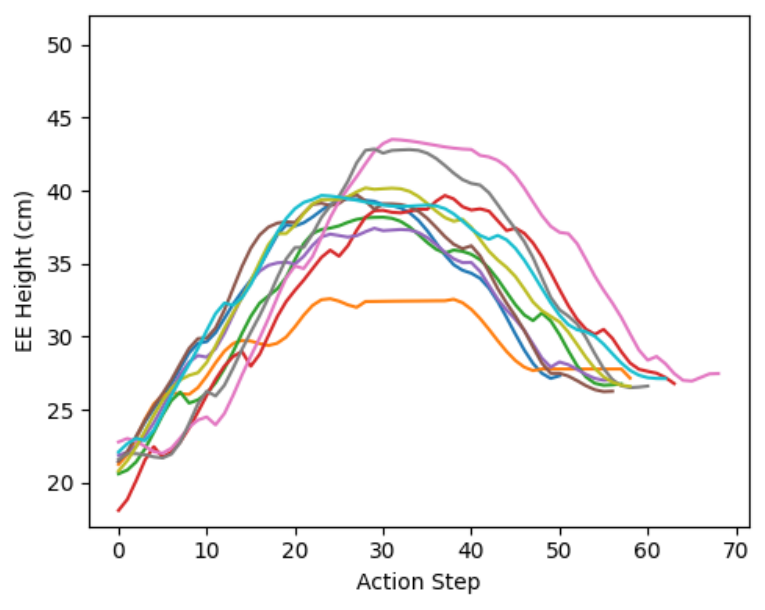}
        \caption{Low Intervention}
        \label{fig:low_trajectories_plot}
    \end{subfigure}
    \hfill
    \begin{subfigure}[t]{0.49\textwidth}
        \centering
        \includegraphics[width=\textwidth]{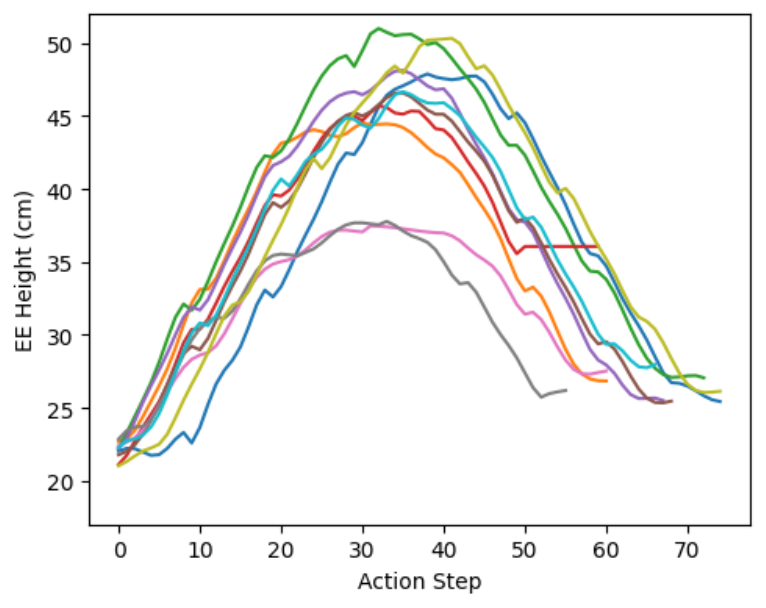}
        \caption{High Intervention}
        \label{fig:high_trajectories_plot}
    \end{subfigure}
    \caption{\textbf{Low/High Transport:} End-effector height from 10 trajectories with low intervention (a) vs. high intervention (b). We plot the segment of the trajectory from when the robot picks up the penguin to when it releases it into the basket. Low interventions have a lower maximum height on average.}
    \label{fig:low_high_trajectories_comparison}
\end{figure}

\begin{figure}[h]
    \centering
    \begin{subfigure}[t]{0.48\textwidth}
        \centering
        \includegraphics[width=\textwidth]{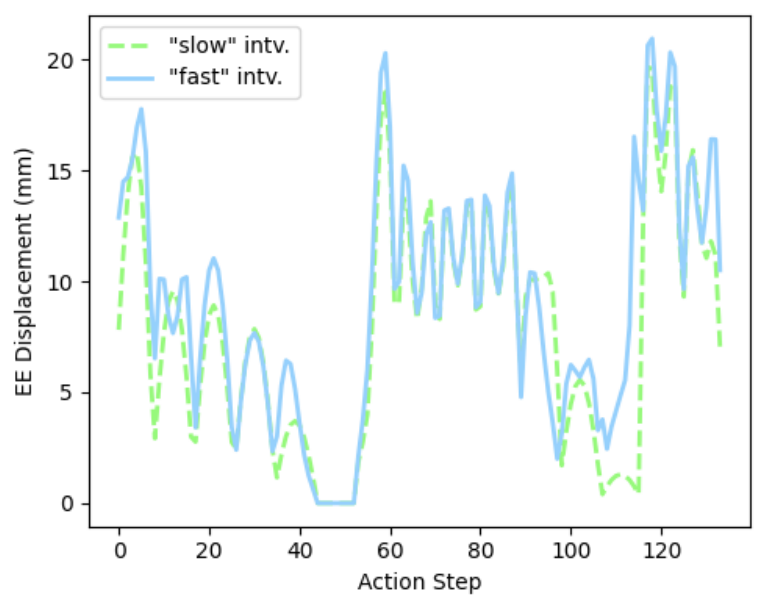}
        \caption{End-effector Displacement}
        \label{fig:slow_fast_trajectories}
    \end{subfigure}
    \hfill
    \begin{subfigure}[t]{0.48\textwidth}
        \centering
        \includegraphics[width=\textwidth]{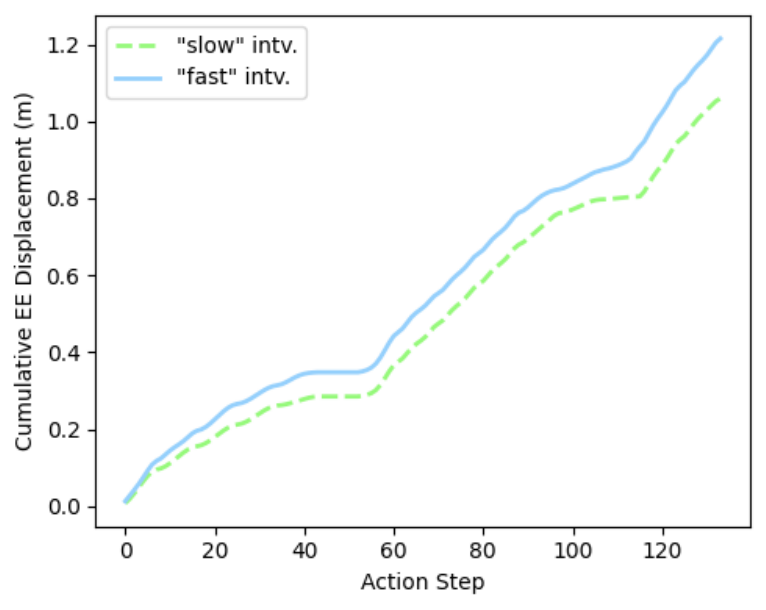}
        \caption{Cumulative End-effector Displacement}
        \label{fig:slow_fast_trajectories_cumulative}
    \end{subfigure}
    \caption{\textbf{Slow/Fast Transport:} End-effector displacement (a) and cumulative end-effector displacement (b) over a single trajectory with slow intervention vs. high intervention. While it may be hard to compare displacement at each action step, the cumulative displacement shows how the slow intervention reduces displacement.}
    \label{fig:slow_fast_trajectories_comparison}
\end{figure}

\end{document}